# Integrating Text and Image Pre-training for Multi-modal Algorithmic Reasoning

Zijian Zhang
Harbin Institute of Technology
Harbin China
zhangzj0318@qq.com

Wei Liu
Harbin Institute of Technology
Harbin China
liuweihit2023@163.com

## Abstract

*In this paper, we present our solution for SMART-101 Challenge of CVPR Multi-modal Algorithmic Reasoning Task 2024. Unlike traditional visual questions and answer tasks, this challenge evaluates abstraction, deduction and generalization ability of neural network in solving visuo-linguistic puzzles designed for specially children in the 6-8 age group. Our model is based on two pre-trained models, dedicated to extract features from text and image respectively. To integrate the features from different modalities, we employed a fusion layer with attention mechanism. We explored different text and image pre-trained models, and fine-tune the integrated classifier on the SMART-101 dataset. Experiment results show that under the data splitting style of puzzle split, our proposed integrated classifier achieves superior performance, verifying the effectiveness of multi-modal pre-trained representations.*

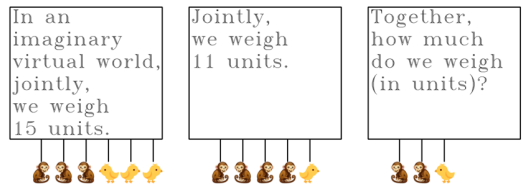

Figure 1: An example in SMART-101, which requires to be solved with the help of equations.

## 1. Introduction

Multimodal visual question answering (VQA) task is an attractive research direction in the field of artificial intelligence in recent years. It integrates technologies from two subfields of computer vision and natural language processing, aiming at building an intelligent system that can understand the content of an image and answer questions based on it. Another important research direction is Multi-modal Algorithmic reasoning (MAR), which aims to guide models to solve complex logic problems based on image context. Due to the similarity of task types, MAR problems often appear in the form of VQA, and MAR tasks can be roughly summed up as a subset of VQA tasks.

In the field of MAR research, there are already some public benchmarks, such as Image riddles[1], PororoQA[3], VLQA[2], and CLEVR[4]. However, while the challenges in these prior works often seem diverse, they tend to be limited to a common setting and also to a specific domain of expertise, allowing existing neural network models to seize on their weaknesses and achieve high accuracy on these datasets. Therefore, the goal of SMART-101[5] is to understand the capabilities and shortcomings of SOTA deep models for visual language inference and provide optimization directions for subsequent work. SMART-101 focuses on investigation the model generalizability to problems that require a wide range of skills, and its data focuses on deeper logical questions, rather than simply answering common-sense math questions. As shown in Figure 1, the problems in SMART-101 are not limited to simple, common-sense questions, but requires specific mathematical methods (equations) and multi-hop reasoning.

To address these challenges, we propose a classification model based on integration of multi-modal pre-trained models. Firstly, we propose a siamese architecture, with two encoders extracting features from text and vision modalities respectively, and then integrated by a fusion layer to produce the classification result. Similar to the cross-attention described in [6], we map the features of the visual tower to the q vector and the features of the text tower to the k and v vectors in the fusion attention mechanism. A linear layer is used to align different representation spaces, and the fused features is fed to a pooling structure to obtain the classification results. Secondly, we explore the utilities of multi-modal pre-trained encoders, which have been pre-trained on massive multi-modal data and is adapted to represent text and image input. Specifically, we explored the utilities of CLIP[14], SigLIP[15] for vision representation, and BERT[8] and DeBERTa[11] for language representation.

We fine-tuned our integrated classifier on the provided training data. Experiment results on the test set showcase that both the integration architecture and the introduction of pre-trained representations improve the accuracy by large margin.

Our contributions can be summarized as follows:
● We propose a Siamese classification architecture for Multimodal Algorithmic reasoning, with different features extracted respectively and fused with attention mechanism.
● We propose the integration of pre-trained encoders from different modalities, to provide better representations for multi-modal input.

## 2. Related Work

### 2.1. Multi-Modal Algorithm Reasoning

CoT is usually used in multi-modal algorithm reasoning. Multimodal-CoT[20] incorporates language (text) and vision (images) modalities into a two-stage framework that separates rationale generation and answer inference. DDCoT[21] proposes a novel prompting that maintains a critical attitude through negative space prompting and incorporates multimodality into reasoning by first dividing the reasoning responsibility of LLMs into reasoning and recognition and then integrating the visual recognition capability of visual models into the joint reasoning process. MC-CoT[22] proposes a self-consistency training strategy that generates multiple rationales and answers, subsequently selecting the most accurate through a voting process.

### 2.2. Pretrained Language Model

BERT[8] pre-trained on large-scale unsupervised text data and then fine-tuned on a specific task. By pre-training on both MLM (Masked Language Model) and NSP (Next Sentence Prediction) tasks, BERT can learn rich contextual information. BERT has achieved significant performance gains on tasks such as text classification, named entity recognition, question answering systems, and natural language inference. Compared with BERT, RoBERTa[9] increases the amount of data training and training time. The NSP task was removed and the mask strategy of dynamically adjusting the Masked Language Model (MLM) is adopted. ALBERT[10] factorizes the large word embedding matrix into the product of two smaller matrices, and the parameters are shared between all layers. Sentence Order Prediction (SOP) is used instead of NSP task. DeBERTa[11] processes content and location information separately by introducing decoding enhancement and disentangled attention mechanism, which can better capture the relationship between words and improve the performance of the model.

### 2.3. Vision Language Pretraining

ViT[13] applies traditional Transformer architectures, originally designed to process sequential data such as text, to the vision domain. ViT first cuts an image into a series of patches, which are then fed into a Transformer model as sequence data. CLIP[14] is a multi-modal pre-trained model proposed by OpenAI that understands both images and text. CLIP is trained by contrastive learning of many images and corresponding description texts. SigLIP[15] proposes a simple pairwise sigmoid loss algorithm for language-image pre-trained based on CLIP. Unlike standard contrastive learning with softmax normalization, the sigmoid loss operates solely on image-text pairs and does not require a global view of the pairwise similarities for normalization.

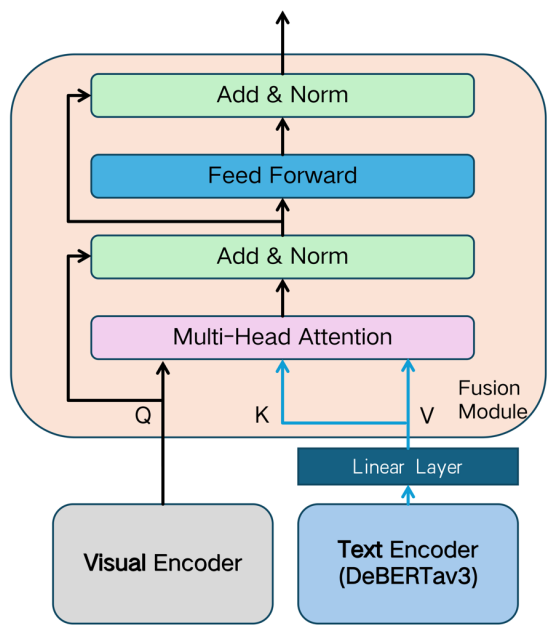

Figure 2: Architecture of the text encoder and fusion module. The output of the text encoder goes through a linear layer to align the text features with the dimensions of the visual features. The fusion module is a multi-head attention mechanism.

## 3. Method

### 3.1. Siamese Classification Model

The model architecture is roughly divided into four parts: visual encoder, text encoder, fusion module, and pooling module.

The overall architecture of the model is shown in Figure 3. The image data is passed through a vision encoder to obtain an image representation. Text data is passed through a text encoder to obtain a text representation. A linear layer is used to align the semantic space with the visual space. The two are input into the fusion module. The architecture of the fusion module is shown in Figure 2. Fusion Module is a Transformer block. We map the features of the visual tower to the q vector and the features of the text tower to the k and v vectors in the fusion attention mechanism. The output of the fusion module goes through the pooling module to obtain the image-text joint feature representation for the classification task.

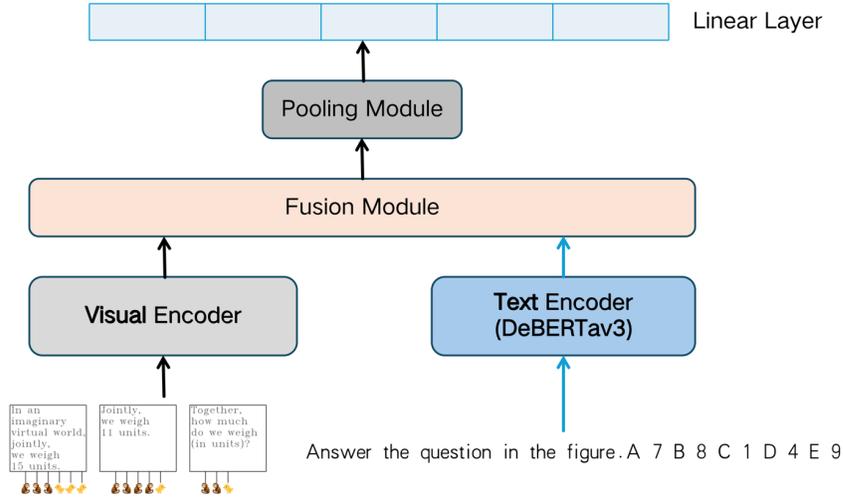

Figure 3: The overall achitecture of the model. The image data and text data are passed through the image encoder and text encoder respectively,and the output is the passed through the fusion module to obtain fusion features for classification tasks.

In the data preprocessing part, we use a preprocessor to process the image data into the input format required by the model. For text data, we concatenate questions with options and construct the text format shown in Figure 3. After obtaining the fused feature output, a linear layer is used to implement a five-class classification model.

### 3.2. Multi-modal Pre-trained Encoder

Based on the architecture introduced in 3.1, we optimize it from different perspectives. These include replacing the text encoder, replacing the visual encoder, and changing the alignment between the visual space and the semantic space.

CLIP is pre-trained by contrastive learning. We adopt the pre-training strategy of CLIP and fine-tune a CLIP-based image-text matching model on SMART-101 dataset as our baseline. Clip-vit-large-patch14 uses the training strategy of CLIP, and ViT-L-14 serves as the visual encoder. We try to employ it as visual encoder of the architecture shown in Figure 3. ViT-SO400M-14-SigLIP-384 is a multi-modal pre-trained model trained on the WebLI[23] dataset using the training strategy of SigLIP. It has excellent performance on many benchmarks. We try to apply its visual encoder and text encoder to the architecture shown in Figure 3.

BERT has achieved significant performance gains on multiple natural language processing tasks by pre-training on large-scale unsupervised text data through a bidirectional Transformer architecture and then fine-tuning on specific tasks. Based on BERT, DeBERTa-v3-large further improves the performance and generalization ability of the model by introducing disentangled attention mechanism, relative position encoding, improved pre-training tasks, and a larger model size. DeBERTa-v3-large has demonstrated stronger performance on multiple natural language processing tasks. We also try to employ them as the text encoder to the architecture shown in Figure 3.

## 4. Experiments

### 4.1. Setup

SMART-101 has 101 root puzzles and 2000 image-text pairs are generated from each root puzzle individually. Under the data splitting mode of Puzzle Split (PS), the root puzzles are split into 77-3-21 (train-val-test)[5]. In this setting, a total of 154000 image-text pairs is used for training, 6000 image-text pairs for validation, and 42000 image-text pairs for training. An example of SMART-101 is shown in Figure 1. From this SMART-101 input and output format, we can find that the accuracy rate is very appropriate for evaluating the performance of the model.

After we find the best method locally, we train on the whole dataset to get the best result on the remote private test set.

### 4.2. Results

The results of the different methods are shown in Table 1. As described in 3, we experimented with different combinations of visual encoders and text encoders. In addition, we experimented with two different alignments, semantic space to visual space alignment and visual space to semantic space alignment. We use a similar pooling approach to BERT. In particular, the method in the last row of Table 1,

| Vision | Text | Align | Pool | L_acc | R_acc |
|---|---|---|---|---|---|
| CLIP.v | CLIP.t | NA | NA | 22.23 | 18 |
| CLIP_ViT_L_224 | BERT | T_to_I | BERT | 25.52 | NA |
| CLIP_ViT_L_336 | BERT | T_to_I | BERT | 24.63 | NA |
| CLIP_ViT_L_224 | BERT | I_to_T | BERT | 24.99 | NA |
| CLIP_ViT_L_224 | DeBERTa | T_to_I | BERT | 25.89 | NA |
| SigLIP.v | SigLIP.t | T_to_I | BERT | 20.22 | NA |
| SigLIP.v | DeBERTa | T_to_I | Attn-pool | 26.14 | 28 |

Table 1: Accuracy of different methods on local test set (L_acc) and remote private test set (R_acc). Align refers to the alignment of visual space and semantic space. T_to_I refer to the mapping from semantic space to visual space. CLIP.v refers to the visual encoder of the CLIP model, and CLIP.t refers to the text encoder of the CLIP model. DeBERTa refers to the DeBERTa-v3-large. SigLIP refers to the model ViT-SO400M-14-SigLIP-384.

we use the attn-pool pooling structure of ViT-SO400M-14-SigLIP-384.

Four conclusions can be drawn from the experimental results. Firstly, the task is difficult. The various methods used in this paper only achieve the accuracy close to random selection. Secondly, using a stronger pre-trained model can improve accuracy. SigLIP outperforms CLIP. DeBERTa outperforms BERT. Thirdly, text-only pre-trained models such as DeBERTa perform better compared to pre-trained text encoders such as SigLIP.t. Finally, the matching method based on CLIP does not perform well on such task that requires a deep understanding, while the method based on classification performs better on it.

## 5. Conclusion

In this paper, we propose an integrated classifier for multi-modal algorithm reasoning. Our model is based on two pre-trained models, dedicated to extract features from text and image respectively. To integrate the features from different modalities, we employed a fusion layer with attention mechanism. Experiment results verifies the effectiveness of multi-modal pre-trained representations.

In the future, we will apply our fusion mechanism to generative models, such as T5[19]. This approach may be able to build MLLM with better understanding capabilities.